\documentclass[conference]{IEEEtran}

\usepackage{cite}
\usepackage{amsmath,amssymb,amsfonts}

\usepackage{graphicx}
\usepackage{textcomp}
\usepackage{xcolor}
\usepackage{multirow}
\usepackage[absolute,overlay]{textpos}
\usepackage{algpseudocode}
\usepackage{algorithm}
\usepackage{pifont}
\usepackage{hyperref}

\def\BibTeX{{\rm B\kern-.05em{\sc i\kern-.025em b}\kern-.08em
    T\kern-.1667em\lower.7ex\hbox{E}\kern-.125emX}}
\begin{document}

\begin{textblock*}{5cm}(\dimexpr\paperwidth-5cm-1cm\relax, 1cm)
    \raggedleft \fontsize{10pt}{12pt}\selectfont Special Session
\end{textblock*}

\title{BearingNAS: Obtaining In-Sensor Intelligent Fault Diagnosis Systems for Bearings Using a Laptop}

\author{
    \IEEEauthorblockN{Andrea Mattia Garavagno\IEEEauthorrefmark{1}\IEEEauthorrefmark{2}, Edoardo Ragusa\IEEEauthorrefmark{1}, Paolo Gastaldo\IEEEauthorrefmark{1}, Antonio Frisoli\IEEEauthorrefmark{2} and Rodolfo Zunino\IEEEauthorrefmark{1}}
    \IEEEauthorblockA{\IEEEauthorrefmark{1}Department of Naval, Electrical, Electronic and Telecommunications Engineering (DITEN), University of Genoa, Italy\\
    Emails: andreamattia.garavagno@edu.unige.it, \{edoardo.ragusa, paolo.gastaldo, rodolfo.zunino\}@unige.it}
    \IEEEauthorblockA{\IEEEauthorrefmark{2}Institute of Mechanical Intelligence, Sant'Anna School of Advanced Studies, Pisa, Italy\\
    Emails: \{andreamattia.garavagno, antonio.frisoli\}@santannapisa.it
    }
    \IEEEauthorblockA{Corresponding Author: Andrea Mattia Garavagno}
}

\maketitle

\begin{abstract}
This paper introduces BearingNAS, a Hardware-Aware Neural Architecture Search (HW-NAS) framework designed to shift the intelligence directly onto the sensor die via in-sensor processing. BearingNAS frames the search as a constrained optimization problem targeting extreme micro-budgets (4 to 8 kiB of RAM and 16 to 32 kiB of Flash). To eliminate the reliance on expensive discrete GPUs, we propose a lightweight, derivative-free search strategy paired with a single data-flow search space that leverages a decaying kernel growth formulation to prevent parameter explosion. We evaluate our framework on the Case Western Reserve University (CWRU) bearing benchmark, optimizing architectures for three STMicroelectronics targets: two commodity microcontrollers and the LSM6DSO16IS Intelligent Sensor Processing Unit (ISPU). Running entirely on a laptop CPU, the search converges in less than an hour. The resulting best in-sensor architecture achieves a highly competitive diagnostic accuracy of 99.50\% on the ISPU. These results demonstrate the viability of shifting the machine learning workload inside the sensor package, enabling low-cost, production-scale bearing fault diagnosis.
\end{abstract}


\section{Introduction}
Bearings are essential mechanical components that critically impact the reliability of multiple industrial sectors, including railways \cite{yan2022survey}, production machinery \cite{miao2021vibration}, wind power \cite{gbashi2024academic}, robotics \cite{xue2026cross}, aerospace \cite{kumar2023bearings}, nuclear \cite{zhu2026progress}, and petroleum \cite{feng2025advanced}. Consequently, identifying bearing faults is paramount to ensuring continuous operations; historically, approximately 40–50\% of rotating equipment failures are attributed to bearing degradation \cite{he2025dcaggcn}. When a bearing fails, the ramifications can cascade through the entire system, resulting in severe economic losses or even catastrophic casualties \cite{cerrada2018review}.

To mitigate these risks, researchers have long focused on both the robust design and the automated identification of bearing faults \cite{hamrock1983rolling}. Recently, state-of-the-art machine learning techniques, ranging from Support Vector Machines (SVM) \cite{widodo2009fault} and Long Short-Term Memory (LSTM) networks \cite{gao2021novel} to Convolutional Neural Networks (CNN) \cite{chen2020rolling}, Transformers \cite{ding2022novel}, and Large Language Models (LLM) \cite{li2025fd}, have been widely adopted for fault diagnosis. However, the vast majority of these models rely on powerful, power-hungry external computational units, rendering large-scale, cost-effective deployment at the edge impractical \cite{chen2023deep}. Conversely, existing fault diagnosis algorithms optimized for edge deployment often demand specialized, highly manual engineering skills to be deployed into real-world industrial scenarios \cite{yoo2023lite}.

To bridge this gap, we introduce BearingNAS, an end-to-end framework designed to automatically generate state-of-the-art Intelligent Fault Diagnosis Systems (IFDS) tailored for commodity microcontrollers and resource-constrained Intelligent Sensor Processing Units (ISPU) embedded within modern Inertial Measurement Units (IMU). By leveraging the compact form factor, high energy efficiency, and sub-dollar cost of these target electronics, our solution democratizes access to IFDS, enabling the ubiquitous integration of localized fault-diagnosis capabilities across massive production scales.

Code is available at \url{https://github.com/AndreaMattiaGaravagno/BearingNAS}.

\section{Related Works}
Early iterations of Intelligent Fault Diagnosis Systems (IFDS) relied heavily on manual feature extraction from vibration or acoustic signals, followed by traditional machine learning classifiers such as Support Vector Machines (SVM) \cite{widodo2009fault}. While effective under constrained laboratory settings, these methods struggled to cope with the non-linear, non-stationary noise characteristic of real-world industrial environments.

To overcome these limitations, deep learning (DL) has emerged as the dominant paradigm. Deep architectures inherently learn hierarchical feature representations directly from raw or minimally processed sensor data. Long Short-Term Memory (LSTM) networks \cite{gao2021novel} have been widely deployed to capture temporal dependencies in sequential signal streams, while Convolutional Neural Networks (CNN) exploit spatial relationships when signals are converted into time-frequency spectrograms \cite{santamato2024leveraging} or processed directly as time-varying temporal sequences \cite{ragusa2024compression}. More recently, the field has transitioned toward high-capacity architectures. Transformers \cite{ding2022novel} leverage self-attention mechanisms to model long-range correlations in sensor data, and contemporary frameworks have even begun evaluating Large Language Models (LLM) \cite{li2025fd} for zero-shot anomaly classification. However, a systemic drawback across this lineage of DL research is an exclusive focus on maximizing classification accuracy, which comes at the expense of excessive and unsustainable computational overhead.

Driven by the need for low latency, data privacy, and reduced communication bandwidth, transferring IFDS capabilities from centralized cloud servers to the industrial edge is emerging as a critical focus \cite{yoo2023lite, vitolo2022low}. This paradigm shift is widely known as TinyML, where machine learning (ML) inference is implemented natively on resource-constrained embedded devices. 

However, deploying state-of-the-art ML algorithms on such restricted hardware requires specialized hardware co-design skills that differ significantly from those typically found in the fault diagnosis field \cite{tognocchi2026automated}. Hardware-Aware Neural Architecture Search (HW-NAS) has emerged as a leading solution to automate the design of ML models tailored to tight resource constraints \cite{benmeziane2021comprehensive}. Early HW-NAS frameworks were computationally prohibitive, requiring up to 40,000 Graphics Processing Unit (GPU) hours to identify an optimal solution \cite{tan2019mnasnet}. Subsequent iterations successfully reduced this search cost to 300 GPU hours \cite{lin2020mcunet} and eventually to just 3 GPU hours \cite{garavagno2024colabnas}, ultimately eliminating the need for GPU acceleration entirely \cite{garavagno2024affordable} and enabling search directly on edge devices \cite{garavagno2025searching}. 

Despite these advancements, none of these HW-NAS frameworks explicitly account for the peculiarities of in-sensor computing. For instance, ColabNAS \cite{garavagno2024colabnas} focuses primarily on image inputs; while images are a widely adopted input format for analyzing vibrations, processing them exceeds the strict resource limits of ISPUs. Similarly, although NanoNAS \cite{garavagno2024affordable, garavagno2025searching} achieves state-of-the-art results on the CWRU dataset, the hardware resources required by its resulting architectures are larger than the resources available on the ISPU.

To bridge the gap between the expertise required to design bearing fault diagnosis systems and the specialized knowledge needed for TinyML implementation, we present BearingNAS. Our approach introduces an HW-NAS framework optimized to deliver In-Sensor Intelligent Fault Diagnosis Systems (IS-IFDS). By operating efficiently on a commodity laptop, BearingNAS eliminates heavy infrastructure barriers while delivering diagnostic performance comparable to state-of-the-art, cloud-tethered IFDS for bearings.

\section{BearingNAS}
BearingNAS is an HW-NAS designed to run on laptops without a discrete GPU. It targets resource-constrained microcontrollers or ISPUs having between 4 and 8 kiB of RAM and 16 to 32 kiB of Flash, such as those reported in Table \ref{tab:sensor_nodes}. 

\subsection{Problem Statement}

\begin{equation} \label{eq:problem}
\begin{aligned}
\begin{cases}
   \hfil  \underset{A}{\mathrm{argmin}} f(A)\\
   \hfil \phi_{RAM}(A) \leq \xi_{RAM} , \phi_{Flash}(A) \leq \xi_{Flash} \\
   \hfil \phi_{MAC}(A) \leq \xi_{MAC} \\
   \hfil \xi_{RAM}, \xi_{Flash}, \xi_{MAC}  > 0
\end{cases}
\end{aligned}
\end{equation}

The main objective is to find the best neural architecture fitting the resource constraints imposed by the target hardware, which can be cast as an optimization problem. In the case of microcontrollers, RAM and Flash capacities dictate the feasibility of a generic candidate architecture. Conversely, the number of Multiply-and-Accumulate (MAC) instructions rules the latency. Therefore, RAM, Flash, and MAC have been adopted as the constraints of the optimization problem, as shown in Equation \ref{eq:problem}. In detail, the functions $\phi_{RAM}$, $\phi_{Flash}$, and $\phi_{MAC}$ return the RAM, Flash, and MAC footprints of a candidate architecture $A$; while $\xi_{RAM}$, $\xi_{Flash}$, and $\xi_{MAC}$ define the upper bounds imposed by the chosen target hardware. Among the feasible solutions, we select the architecture that minimizes the validation loss during training, which is measured by the function $f$ in Equation \ref{eq:problem}. 

\subsection{Search Space}
The search space defines how a generic candidate architecture $A$ is built. Given the tight resource constraints of the target hardware, complex search spaces comprising directed acyclic graphs (DAGs) or residual connections have been avoided to reduce RAM and Flash usage. Instead, a search space with a single data-flow has been adopted. Candidates are built stacking convolutional, pooling, and batch-normalization layers. Every candidate starts with a convolutional layer featuring $k$ kernels of size $1 \times 3$. Then, $c$ cells, each composed of a $1 \times 2$ max pooling layer, a batch-normalization layer, and a convolutional layer featuring $n$ kernels of size $1 \times 3$, are stacked. Finally, a global average pooling layer compresses the extracted features and a dense layer with softmax activation classifies them. The number of kernels $n$ adopted by the convolutional layer of a generic cell is defined by the following equation: $n_{c} = n_{c - 1} + \lfloor2^{1-c} n_{c - 1}\rfloor$, where $n_{0} = k$. This decaying growth formulation ensures the number of kernels expands at a progressively diminishing rate, deliberately preventing the parameter and activation memory explosion that typically bottlenecks deep networks on resource-constrained microcontrollers. Every convolutional layer adopts the hardware-friendly ReLU activation. The result is a bi-dimensional search space where candidates can be unequivocally identified by the tuple $(k,c)$. 

\subsection{Search Strategy}
The search strategy defines how to find the solution of the optimization problem. To accommodate the computational capabilities of laptops lacking a discrete GPU, a custom derivative-free search strategy, inspired by NanoNAS \cite{garavagno2025searching}, has been adopted. It is composed of two nested loops, an outer one (Algorithm \ref{alg}, line \ref{alg:outer_level}) exploring the $k$ axis of the search space and an inner one (Algorithm \ref{alg}, line \ref{alg:inner_level}) exploring the $c$ axis. Starting from zero, the inner loop stacks cells upon the base convolutional layer having $k$ kernels, one after another, to identify the best feasible depth for the current value of k. Then, the outer loop proposes a new number of kernels $k$ of the base convolutional layer according to the equation in line \ref{alg:k} of Algorithm \ref{alg}. 

The exhaustive exploration of the $c$-axis is a key difference from NanoNAS \cite{garavagno2025searching}, where the search was truncated upon the first performance decrease. This early stopping mechanism prevented the search strategy from exploring deeper architectures at low $k$ values due to training noise, thereby hindering the HW-NAS from converging to more performant, lightweight solutions. Furthermore, we employ an objective function based on validation loss rather than validation accuracy; this accounts for the data scarcity typical of intelligent fault diagnosis, which caused the original algorithm to terminate prematurely upon finding a perfect score on the validation subset adopted during the search procedure.

The search procedure starts from the candidate $(1,0)$, the smallest one, and alternates the exploration of the two directions of the search space. The value $k$ is gradually increased as long as a better candidate is found on the $c$ axis. When the first decrease in the objective function is found between two movements on the $k$ axis, the optional increment reduction mechanism is triggered (Algorithm \ref{alg}, line \ref{alg:dec}) to check if a smaller increment can provide a better solution. When the variable increment becomes zero, the search stops and the algorithm returns the best solution found. 

\begin{algorithm}
\begin{algorithmic}[1]
\Procedure{search strategy}{}
\State $k\gets 1$, $c\gets 0$ \Comment{Starting point}
\State $\beta\gets 0$, $\gamma\gets 0$
\State $\bar{k}\gets k$
\Repeat \label{alg:outer_level} \Comment{Outer loop}
\State $\bar{c}\gets 0$, $c^{*}\gets 0$
\While{$(\bar{k},\bar{c}+1)$ is feasible} \label{alg:inner_level} \Comment{Inner loop}
\If{$f(\bar{k},\bar{c}+1) < f(\bar{k},c^{*})$}
\State $c^{*}\gets \bar{c}+1$  \Comment{Update of best c}
\EndIf
\State $\bar{c}\gets \bar{c} + 1$
\EndWhile
\If{$f(\bar{k},c^{*}) < f(k,c)$} 
\State $k\gets \bar{k}$, $c\gets c^{*}$ \Comment{Candidate confirmation}
\Else
\State $\gamma\gets 1$ \Comment{Start k increment reduction}
\EndIf
\State $\beta\gets \beta + \gamma$ \label{alg:dec} \Comment{Optional k increment reduction}
\State $\bar{k}\gets k + \lfloor 2^{-\beta}k \rfloor$ \label{alg:k} \Comment{Variable increment of k}
\Until{$\lfloor 2^{-\beta}k \rfloor = 0$} \Comment{Stopping criterion}
\EndProcedure
\end{algorithmic}
\caption{Proposed search strategy.}\label{alg}
\end{algorithm}

\section{Experimental Setup}
To evaluate BearingNAS, we utilized the Case Western Reserve University (CWRU) bearing dataset, a widely adopted benchmark for fault diagnosis in industrial condition monitoring \cite{chen2023deep}. Provided by the CWRU Bearing Data Center, this dataset consists of multivariate time-series acceleration data captured from both the drive-end and fan-end of a 2-horsepower electric motor under varying speeds and loads.

The data were generated using a SpectraQuest machinery fault simulator fitted with SKF 6205 deep groove ball bearings. Faults were artificially induced using electro-discharge machining at three severity levels: 0.007, 0.014, and 0.021 inches. Accelerometers (PCB 353B33, 100 mV/g sensitivity) recorded the data, which was downsampled to 12 kHz across four operating loads ranging from 0 to 3 horsepower (1797 down to 1690 rpm).

Our evaluation focuses on the 10-class configuration using the drive-end sensor at the 0 hp baseline. This encompasses one healthy state and nine fault scenarios: ball and inner raceway faults at 0.007, 0.014, and 0.021 inches; and outer raceway faults oriented at 3, 6, and 12 o'clock at 0.007 inches. All were selected at 1797 rpm (i.e., at 0 hp). The dataset was balanced by selecting 200 non-overlapping segments per class, consisting of 512 samples each, effectively capturing the characteristic impulsive patterns and specific fault frequencies. We employ a strict chronological 70/10/20 split for training, validation, and testing to prevent time-series data leakage and ensure our hardware-aware architecture optimization generalizes effectively to unseen conditions. Results from the literature may use different train/test protocols.

During the search process, we evaluate each candidate architecture against the validation set after 100 epochs of training with a batch size of 16 and a learning rate of 0.001 using the Adam optimizer. We perform all NAS computations on a laptop equipped with an Intel(R) Core(TM) Ultra 7 155H and 32 GB of RAM running at 7467 MT/s.

To benchmark the hardware-awareness of BearingNAS, we target ultra-low-power devices ranging from 4 kiB of RAM and 16 kiB of Flash up to 8 kiB of RAM and 32 kiB of Flash. Specifically, we target two microcontrollers, the STM32F030F4P6 and the STM32C011F6P6, and one ISPU, the LSM6DSO16IS, whose specifications are detailed in Table \ref{tab:sensor_nodes}. To set the MAC upper bound we multiply the CoreMark score of each platform by a factor of $10^4$ \cite{garavagno2025searching}. 
As no standardized CoreMark benchmark is available for the ISPU, we adopt the same MAC bound used for the STM32C011F6P6 as a practical proxy.

To measure the Flash and RAM footprint of candidate architectures, we utilize X-CUBE-AI 9.1.0 with the TFLite Micro runtime, whereas MAC operations are computed analytically by BearingNAS. Finally, we measure the latency of the resulting neural architectures on the LSM6DSO16IS using the ST Edge AI Developer Cloud with ST Edge AI Core 4.0.1.

\begin{table}[]
    \centering
    \caption{Hardware Targets}
    \label{tab:sensor_nodes}
    \begin{tabular}{c c c c c c}
       \multirow{2}{*}{Target}        & \multirow{2}{*}{Core}                    & Frequency & CoreMark & RAM & Flash   \\ 
                     &                          & [MHz]     & Score    & [kiB] & [kiB] \\ \hline
       STM32F030F4P6 & M0                       & 48        & 106      & 4     & 16     \\
       STM32C011F6P6 & M0+                      & 48        & 114      & 6     & 32    \\
       LSM6DSO16IS   & ISPU                     & 5         & n.a.     & 8     & 32    \\ \hline
    \end{tabular}
\end{table}




\section{Results and Discussion}

\subsection{Hardware-Aware Architecture Scaling}

\begin{table}[!h]
    \centering
    \caption{Hardware Awareness of BearingNAS}
    \label{tab:sensor_nodes_awareness}
    \begin{tabular}{c c c c c c}
       \multirow{2}{*}{Target} & MAC & RAM   & Flash   & Latency & Accuracy \\ 
                               & [k]  & [kiB] & [kiB]   & [ms]    & [\%]   \\ \hline
       STM32F030F4P6           & 7.7  & 4     & 3.98    & 157.6   & 85.25 \\ 
       STM32C011F6P6           & 70.8 & 6     & 8.99    & 504.5   & 98.75 \\ 
       LSM6DSO16IS             & 97.9 & 7     & 9.64    & 624.6   & 99.50 \\ \hline
    \end{tabular} \\
\end{table}

Table \ref{tab:sensor_nodes_awareness} compares the memory footprint, computational requirements, latency, and accuracy of the resulting BearingNAS models when targeting the three sensor nodes presented in Table \ref{tab:sensor_nodes}. All latency values were measured on the LSM6DSO16IS using the architectures generated for each hardware constraint. Table \ref{tab:sensor_nodes_awareness} demonstrates the ability of BearingNAS to dynamically tailor neural architectures to the specific resource bounds of ultra-low-power target devices. As the hardware constraints relax, the algorithm naturally exploits the available memory to generate deeper or wider models. For the highly constrained STM32F030F4P6 (limited to 4 kiB RAM), the search yields a minimalist architecture requiring only 7.7k MAC operations and 3.98 kiB of Flash, achieving a baseline accuracy of 85.25\%.

When targeting the slightly more capable STM32C011F6P6, the accuracy jumps significantly by 13.5 percentage points (reaching 98.75\%). This performance leap is achieved by utilizing 6 kiB of RAM and expanding the Flash footprint to 8.99 kiB, which allows the network to extract more complex features. Finally, targeting the LSM6DSO16IS ISPU allows the network to reach an accuracy of 99.5\%, utilizing only 7 kiB RAM and 9.64 kiB Flash.

\subsection{The Accuracy-Latency Trade-off}
The results in Table \ref{tab:sensor_nodes_awareness} also highlight the inherent trade-off between diagnostic performance and inference latency on edge devices. The lightweight model generated for the STM32F030F4P6 executes in just 157.6 ms. In contrast, the highly accurate model generated for the LSM6DSO16IS requires 97.9k MAC operations, resulting in a latency of 624.6 ms. By casting MACs as a constraint in the optimization problem (as defined in Equation 1), BearingNAS provides engineers with the flexibility to dictate this trade-off, ensuring that the resulting latency remains within the real-time requirements of the specific industrial condition monitoring application.

\subsection{Search Cost and Laptop-Feasibility}
A primary objective of BearingNAS is to democratize Hardware-Aware NAS by eliminating the reliance on discrete GPUs. Table \ref{tab:search_cost} validates this approach, detailing the search execution time on a laptop CPU. The search cost scales proportionally with the target hardware's resource budget. Because the search space is bounded by hardware constraints, searches for highly constrained targets terminate faster; the search for the STM32F030F4P6 completed in just 4 minutes and 36 seconds.

\begin{table}[!h]
    \centering
    \caption{BearingNAS Search Cost For Each Target}
    \label{tab:search_cost}
    \begin{tabular}{c c}
       \multirow{2}{*}{Target} & Search Cost  \\ 
                               & [mm]:[ss]    \\ \hline
       STM32F030F4P6           & 04:36        \\ 
       STM32C011F6P6           & 24:59        \\ 
       LSM6DSO16IS             & 51:39        \\ \hline
    \end{tabular}
\end{table}

Even for the least constrained target in our evaluation (the LSM6DSO16IS), the search algorithm converged on the best architecture found in under 52 minutes. This rapid turnaround time shows that the proposed single data-flow search space, combined with the derivative-free variable-increment search strategy, is highly efficient and perfectly suited for laptops.

\subsection{Comparison with the State of the Art}
\begin{table}[!h]
    \centering
    \caption{Comparison with representative methods reported on the CWRU dataset}
    \label{tab:sota_comparison}
    \begin{tabular}{c c c c}
        Technique                               & Accuracy & Near-Sensor & In-Sensor \\ \hline
        SVM \cite{wen2017new}                   & 89.20    & \ding{51}   & \ding{55} \\
        LSTM \cite{wang2022convolutional}        & 95.55    & \ding{51}   & \ding{55} \\
        1D-CNN \cite{chen2020improved}           & 99.30    & \ding{51}   & \ding{55} \\
        NanoNAS \cite{garavagno2025searching} & 99.50    & \ding{51}   & \ding{55} \\
        DCLSG \cite{cai2024debiased}             & 99.60    & \ding{51}   & \ding{55} \\ 
        VGG16 \cite{chen2020rolling}             & 99.90    & \ding{55}   & \ding{55} \\        
        InceptionV3 \cite{chen2020rolling}       & 99.94    & \ding{55}   & \ding{55} \\
        LiteCNN \cite{yoo2023lite}               & 99.95    & \ding{51}   & \ding{55} \\ 
        ResNet-50 \cite{chen2020rolling}         & 99.97    & \ding{55}   & \ding{55} \\ \hline
        Proposal                                & 99.50    & \ding{51}   & \ding{51} \\ 
    \end{tabular}
\end{table}

Table \ref{tab:sota_comparison} compares the best architecture generated by BearingNAS against several state-of-the-art methodologies evaluated on the CWRU bearing dataset. Near-sensor capability indicates that the method can reasonably execute on a microcontroller-class edge processor according to the architecture complexity reported by the authors, whereas in-sensor capability denotes demonstrated execution within the sensor package itself. The proposed solution achieves a highly competitive accuracy of 99.50\%. While deep, over-parameterized architectures such as VGG16 \cite{chen2020rolling}, InceptionV3 \cite{chen2020rolling}, and ResNet-50 \cite{chen2020rolling} marginally outperform our proposal (peaking at 99.97\%), they rely on massive computational and memory resources. Consequently, they completely lack both near-sensor and in-sensor execution capabilities, rendering them unsuitable for real-time, ultra-low-power edge deployments in industrial condition monitoring.

More lightweight approaches, such as 1D-CNN \cite{chen2020improved}, DCLSG \cite{cai2024debiased}, and LiteCNN \cite{yoo2023lite}, successfully bridge the gap to edge computing, enabling near-sensor execution (typically hosted on a microcontroller adjacent to the sensor). These models achieve accuracies ranging from 99.30\% to 99.95\%. However, they still require the raw, high-frequency vibration data to be continuously transmitted over digital interfaces (e.g., I2C or SPI) from the sensor to the host MCU, incurring significant power and latency overheads. The closest competitor, NanoNAS \cite{garavagno2025searching}, achieves the same test accuracy but with a 1.93 times larger RAM footprint, preventing it from meeting the tight constraints of the LSM6DSO16IS.

The critical differentiator of our proposal lies in its hardware-awareness. Among the methods considered, BearingNAS is the only one for which in-sensor execution has been demonstrated. By enabling true in-sensor processing, the proposed architecture analyzes the data directly on the sensor die. This drastically reduces the communication bottleneck on the embedded bus, as the sensor only needs to transmit the final diagnostic label (healthy or specific fault class) rather than the raw 12 kHz time-series data. BearingNAS demonstrates that the ultimate edge computing paradigm, in-sensor execution, can be achieved without making meaningful sacrifices to diagnostic accuracy.

\section{Conclusion}
In this paper, we introduced BearingNAS, a Hardware-Aware Neural Architecture Search methodology designed to democratize the deployment of deep learning on ultra-low-power edge devices. By casting architecture search as a hardware-constrained optimization problem, specifically bounding RAM, Flash, and MAC operations, BearingNAS dynamically scales convolutional networks to fit tightly constrained targets, ranging from traditional microcontrollers to ISPUs. To ensure accessibility without the need for discrete GPUs, we proposed a computationally efficient, derivative-free search strategy paired with a specialized single data-flow search space that deliberately mitigates memory explosion through a decaying kernel growth formulation.

Evaluated on the CWRU bearing dataset, BearingNAS successfully generated highly accurate diagnostic models across various hardware profiles in under an hour using a laptop CPU. Most notably, our methodology yielded an architecture capable of true in-sensor execution on the STMicroelectronics LSM6DSO16IS, achieving a highly competitive accuracy of 99.50\%. This represents a critical paradigm shift for industrial condition monitoring, enabling low-cost, production-scale bearing fault diagnosis.

Future work will explore the integration of quantization-aware training directly into the BearingNAS search loop to further compress the architectural footprint. Additionally, we aim to expand this hardware-aware methodology beyond localized vibration analysis to accommodate multi-sensor fusion scenarios, paving the way for ubiquitous, self-contained intelligent nodes in predictive maintenance applications.

\bibliographystyle{IEEEtran}
\bibliography{my_bib}
\end{document}